\documentclass[11pt]{article}

\usepackage[preprint]{acl}

\usepackage{times}
\usepackage{latexsym}

\usepackage[T1]{fontenc}

\usepackage[utf8]{inputenc}

\usepackage{microtype}

\usepackage{inconsolata}

\usepackage{graphicx}
\usepackage{enumitem}
\usepackage{graphicx}
\usepackage{amsmath}
\usepackage{makecell}
\usepackage{enumitem}
\usepackage[table,xcdraw]{xcolor}
\usepackage{colortbl}

\usepackage[utf8]{inputenc} 
\usepackage[T1]{fontenc}    
\usepackage{hyperref}       
\usepackage{url}            
\usepackage{booktabs}       
\usepackage{amsfonts}       
\usepackage{nicefrac}       
\usepackage{microtype}      
\usepackage{xurl}
\usepackage{amsmath} 
\usepackage{listings} 

\title{From Chains to Graphs: Self-Structured Reasoning for \\General-Domain LLMs}

\author{
 \textbf{Yingjian Chen\textsuperscript{1}},
 \textbf{Haoran Liu\textsuperscript{2}},
 \textbf{Yinhong Liu\textsuperscript{3}},
 \textbf{Sherry T. Tong\textsuperscript{1}},
\\
 \textbf{Aosong Feng\textsuperscript{4}},
 \textbf{Jinghui Lu\textsuperscript{5}},
 \textbf{Juntao Zhang\textsuperscript{6}},
 \\
 \textbf{Yusuke Iwasawa\textsuperscript{1}},
 \textbf{Yutaka Matsuo\textsuperscript{1}},
 \textbf{Irene Li\textsuperscript{1}}\thanks{Corresponding author}
\\
\\
 \textsuperscript{1}University of Tokyo,
 \textsuperscript{2}Texas A\&M University,
 \textsuperscript{3}University of Cambridge,
 \\
 \textsuperscript{4}Yale University,
 \textsuperscript{5}Xiaomi EV,
 \textsuperscript{6}Henan University.
\\
 \small{
   \textbf{} \href{mailto:irene.li@weblab.t.u-tokyo.ac.jp}{irene.li@weblab.t.u-tokyo.ac.jp}
 }
}

\begin{document}
\maketitle
\begin{abstract}
Large Language Models (LLMs) show strong reasoning ability in open-domain question answering, yet their reasoning processes are typically linear and often logically inconsistent. In contrast, real-world reasoning requires integrating multiple premises and solving subproblems in parallel. Existing methods, such as Chain-of-Thought (CoT), express reasoning in a linear textual form, which may appear coherent but frequently leads to inconsistent conclusions. Recent approaches rely on externally provided graphs and do not explore how LLMs can construct and use their own graph-structured reasoning, particularly in open-domain QA. To fill this gap, we novelly explore graph-structured reasoning of LLMs in general-domain question answering. We propose Self-Graph Reasoning (SGR), a framework that enables LLMs to explicitly represent their reasoning process as a structured graph before producing the final answer. We further construct a graph-structured reasoning dataset that merges multiple candidate reasoning graphs into refined graph structures for model training. Experiments on five QA benchmarks across both general and specialized domains show that SGR consistently improves reasoning consistency and yields a 17.74\% gain over the base model. The LLaMA-3.3-70B model fine-tuned with SGR performs comparably to GPT-4o and surpasses Claude-3.5-Haiku, demonstrating the effectiveness of graph-structured reasoning.\footnote{Our code is available at \url{https://github.com/Yingjian-Chen/SGR-Self-Graph-Reasoning}.}
\end{abstract}

\section{Introduction}
Large Language Models (LLMs)~\cite{hurst2024gpt, dubey2024llama} have exhibited impressive performance on a wide range of open-domain natural language understanding and question-solving tasks~\cite{zhao2023survey, bang2023multitask, yang2024ascle, teleki2025survey, liu2025llms}. In recent years, research on reasoning-oriented LLMs has progressed rapidly, showing that explicit intermediate reasoning can significantly enhance complex inference and provide more interpretable explanations of model decisions~\cite{ke2025survey, patil2025advancing}. Representative examples include Chain-of-Thought (CoT)~\cite{wei2022chain} and dedicated reasoning models~\cite{jaech2024openai, guo2025deepseek}. 

\begin{figure}[t]
\centering
    \includegraphics[width=1.0\linewidth]{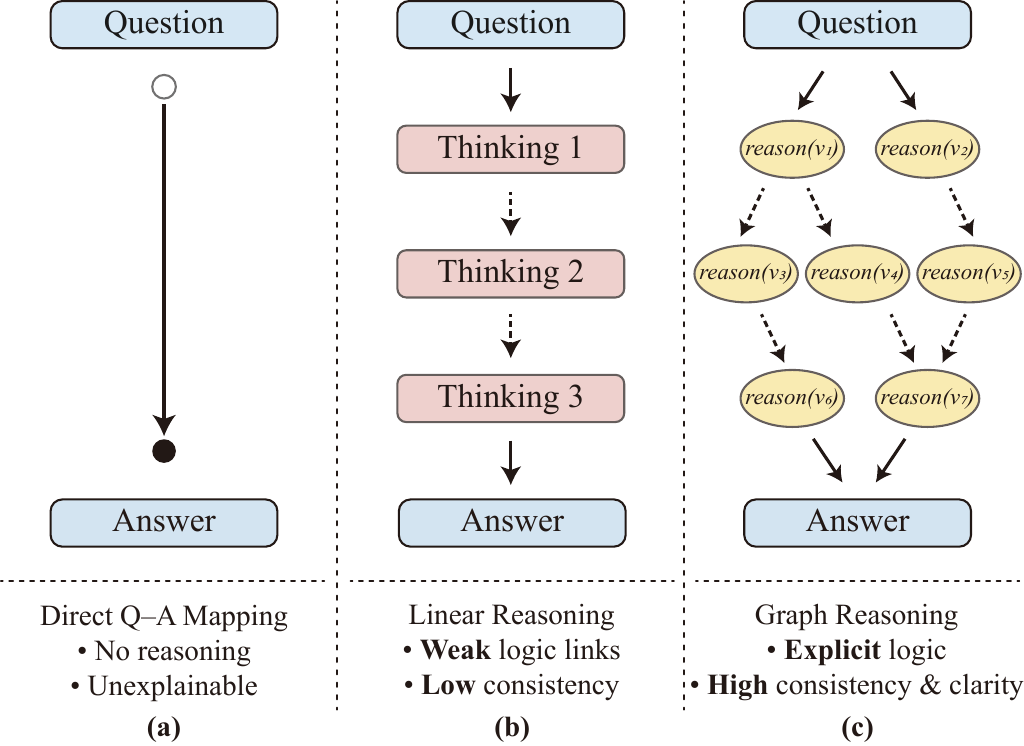}
    \caption{Comparison of reasoning and answering paradigms in general-domain question answering.
    (a) \textbf{Direct Answering} without explicit reasoning.
    (b) \textbf{Linear reasoning} with weak logical alignment between intermediate reasoning and the final answer.
    (c) Our \textbf{Graph-structured reasoning} with explicit logical connections, yielding higher reasoning consistency.
    }
    \label{fig:overview}
\end{figure}

However, current approaches often exhibit inconsistencies between the generated reasoning process and the final answer~\cite{wang-etal-2025-chain, arcuschin2025chain}, especially in general-domain question answering (QA), where the inference lacks a clear and explicit logical structure~\cite{xu2024faithful, lee2025structured}. This makes it difficult to ensure that intermediate steps form a coherent reasoning path toward the final conclusion. Such challenges are less pronounced in specialized domains, such as mathematics and medicine, where reasoning typically follows more formalized and constrained structures. Early paradigms in general-domain question answering focused on direct answer prediction without explicit reasoning (Figure~\ref{fig:overview}, a). Recent advances, such as CoT, introduced intermediate reasoning steps to improve interpretability and performance. However, because the reasoning process in LLMs is represented as a linear textual sequence (Figure~\ref{fig:overview}, b) optimized for next-token prediction rather than for structured logical inference, the reasoning chain may appear superficially coherent yet lead to incorrect answers, or, conversely, flawed reasoning may occasionally yield correct results~\cite{turpin2023languagemodelsdontsay}. Furthermore, unlike explicitly structured reasoning such as knowledge graphs~\cite{chen2020review, tan2025paths, yang2024kg, li2025mkg}, such linear reasoning makes it difficult for LLMs to maintain consistent dependencies between intermediate reasoning steps and the final answer~\cite{patil2025advancing}. These limitations are particularly evident in smaller-scale LLMs. 

To overcome these limitations, we posit that reasoning should extend beyond simple linear sequences. Real-world reasoning often involves parallel sub-problems and the integration of multiple premises, which cannot be adequately represented by a single linear chain. As illustrated in Figure~\ref{fig:overview} (c), graph-structured reasoning enables many-to-one dependencies, allowing multiple independent inferences to be explicitly integrated into a unified conclusion~\cite{yao2024got}. By enforcing explicit parent–child dependencies between reasoning steps, such a structure tightly couples the reasoning process with the final answer, ensuring that each conclusion is grounded in its supporting premises. However, existing graph-based reasoning methods primarily incorporate externally constructed graph structures to guide inference~\cite{jin2024graph, luo2024graph, han2025reasoning}. As a result, whether LLMs can perform explicit self-graph reasoning, particularly in the general-domain, remains largely unexplored.

Building on this insight, we propose \textit{self-graph reasoning} (SGR), a paradigm that compels LLMs to externalize their latent reasoning into an explicitly structured graph before producing a final answer, and we use that as intermediate hints for the model. In SGR, nodes represent reasoning units, while edges encode explicit logical dependencies, forming a structured bridge between the input question and the final prediction. Unlike prior approaches that either rely on linear text-based reasoning or external graph structures, SGR enables LLMs to internalize graph construction as part of the reasoning process itself. Furthermore, we construct a dataset to train models for self-graph reasoning. Given a question, the LLM generates multiple candidate reasoning graphs, which are then aggregated and refined into an optimal reasoning graph, which serves as supervision for training self-graph reasoning. We conduct experiments across five QA benchmarks, covering both general-domain and specialized-domain tasks, demonstrating the effectiveness of our proposed framework. Moreover, we publicly release our constructed graph-reasoning dataset to support future research on structured and interpretable reasoning in LLMs.


In summary, our contributions are: (1) \textbf{Self-Graph Reasoning Method.}     
We introduce a novel reasoning paradigm that enables LLMs to perform structured graph reasoning within the inference process itself, enhancing the transparency and consistency between intermediate reasoning and final answer.
(2) \textbf{Graph-Reasoning Dataset.}
    We construct a general-purpose graph-reasoning dataset of 10K instances that provides explicit structured supervision, enhancing LLMs' capability in graph-based reasoning.
(3) \textbf{Empirical Effectiveness.} 
    We demonstrate the effectiveness of SGR, with a 17.74\% improvement over its base model across five benchmarks while showing consistent effectiveness in specialized domains, including mathematics and medicine.

\section{Related Works}
\noindent\textbf{Chain-of-Thought (CoT) Reasoning.} 
Early language models mainly performed direct answer prediction, where reasoning was implicit~\cite{brown2020language, chowdhery2023palm, li2025implicit}. To address the limited capability in reasoning over complex problems and the lack of interpretability, the Chain-of-Thought (CoT) paradigm was proposed~\cite{wei2022chain, kojima2022large}. CoT encourages models to explicitly generate an intermediate reasoning process before producing the final answer, thereby improving performance on tasks requiring complex logical reasoning~\cite{zhou2022least, zhang2022automaticchainthoughtprompting}. However, despite its effectiveness, CoT reasoning remains essentially linear, modeling reasoning as a single textual sequence optimized for next-token prediction, which can lead to plausible but incorrect reasoning chains~\cite{turpin2023languagemodelsdontsay, lanham2023measuring}. In contrast, our self-graph reasoning represents internal reasoning as a structured topology, where each claim is grounded in its ancestral arguments, resulting in clearer and more coherent reasoning.

\noindent\textbf{Reasoning LLMs.}
Recent studies have shifted toward reasoning LLMs~\cite{patil2025advancing}, which primarily enhance multi-step reasoning through reasoning-oriented training, encouraging models to generate or internally perform reasoning before producing final answers, often via specialized training objectives.
Representative models include OpenAI o1~\cite{jaech2024openai}, DeepSeek-R1~\cite{guo2025deepseek}, and Qwen3-thinking~\cite{yang2025qwen3}. Despite their remarkable performance, these models still rely on a fundamentally linear reasoning process, similar to CoT. As a result, their reasoning remains limited in capturing complex logical dependencies and maintaining consistency between reasoning and final answers~\cite{turpin2023languagemodelsdontsay}, particularly in general-domain QA tasks with intricate logical structures.

\noindent\textbf{Graph-based Reasoning.}
To address the limitations of linear reasoning, recent studies have explored incorporating graph structures into the reasoning process~\cite{luo2023reasoning, sun2023think, jin2024graph, tian2024graph, cao2024graphreason, chen2025graphcheck}. Specifically, Reasoning on Graphs~\cite{luo2023reasoning} generates a reasoning graph to capture logical relations and use it to generate the final answer. MindMap~\cite{wen2024mindmap} retrieves external evidence graphs and performs reasoning based on these graphs. However, these approaches typically use pre-extracted logical graphs or retrieved knowledge graphs, requiring external structures to support reasoning. In contrast, our work is the first to explore self-graph reasoning, where a reasoning LLM autonomously externalizes its internal reasoning process into a structured topological graph before producing the final answer, particularly in general-domain settings.

\begin{figure}[t]
\centering
    \includegraphics[width=1.0\linewidth]{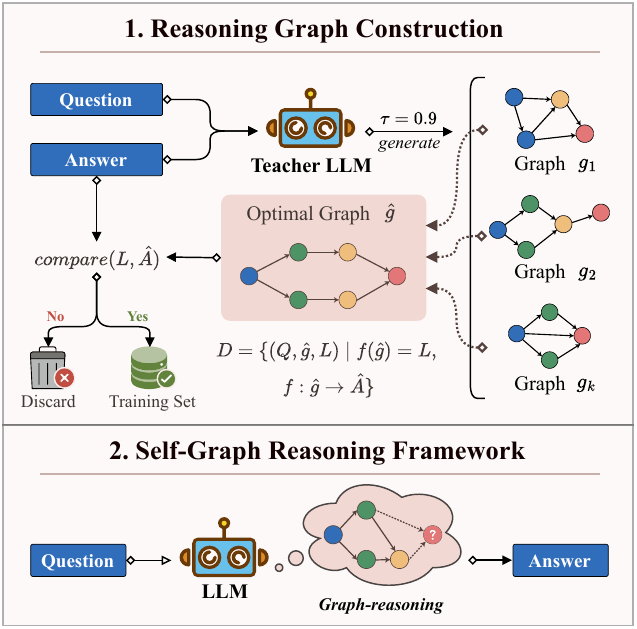}
    \caption{Overview of the proposed Self-Graph Reasoning (SGR).
    }
    \label{fig:pipeline}
\end{figure}

\section{Methods}
In this section, we introduce our proposed Self-Graph Reasoning (SGR), a framework designed to enable LLMs to externalize their latent reasoning process into a structured graph before generating the final answer. By explicitly organizing intermediate reasoning into a graph structure, SGR provides clearer logical constraints and reduces the drift and inconsistency often seen in linear Chain-of-Thought reasoning. The implementation of SGR consists of two main components, as illustrated in Figure~\ref{fig:pipeline}: (1) Reasoning Graph Construction, which constructs a dataset of explicit reasoning graphs capturing the logical process between questions and answers, serving as supervision for model training; and (2) Self-Graph Reasoning Framework, where the model is fine-tuned on the constructed graph data to internalize the ability to perform self-graph reasoning before generating the final answer.

\begin{figure*}[t!]
    \centering
    \includegraphics[width=1\textwidth]{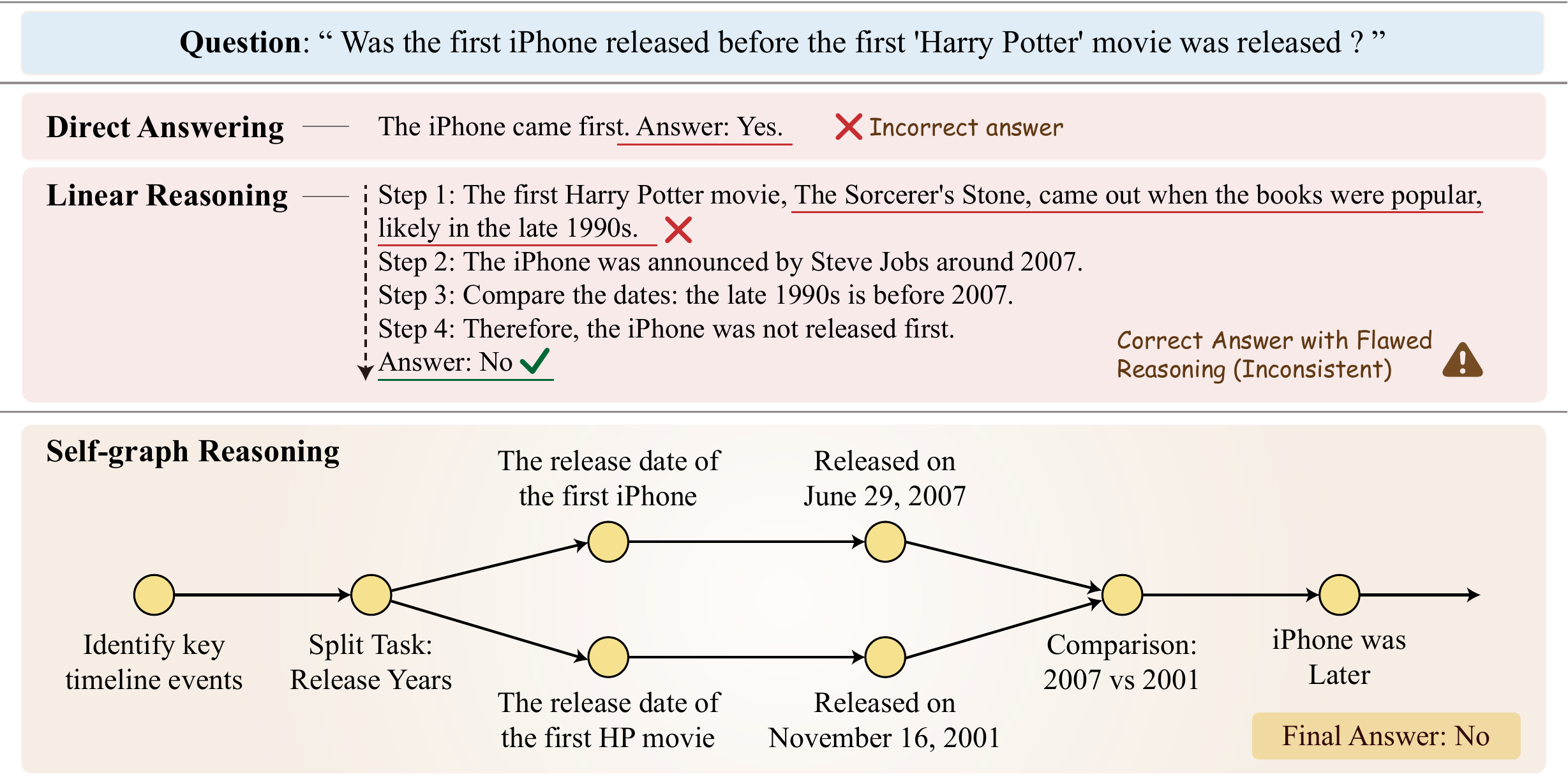}
    \caption{An illustrative example of the Self-Graph Reasoning framework. Our method constructs a structured reasoning graph where each node is explicitly grounded in its predecessors to ensure a logically consistent and clear path to the final answer. For comparison, we also illustrate other paradigms: Direct Answering lacks an explicit thinking process and is prone to errors, while Linear Reasoning often exhibits process-answer inconsistency (e.g., reaching a correct answer through flawed factual steps).}
    \label{fig:sample}
\end{figure*}

\subsection{Reasoning Graph Construction}
\label{reasoning_graph_construction}
To provide the model with high-quality graph reasoning supervision, we transform raw question-answer pairs into structured reasoning graphs.

\noindent\textbf{Diverse Trajectory Exploration.}
To capture a diverse set of reasoning trajectories specific to the general-domain question $Q$, we prompt a teacher LLM (GPT-4o) to explore multiple reasoning paths. General-domain questions often admit multiple reasoning paths, as they can be solved from different starting points or perspectives, each leading to a distinct sequence of logical steps and a valid answer. By setting a higher temperature ($\tau = 0.9$), we encourage stochasticity and diversity in the reasoning process. Formally, for each $Q$, we sample $k$ independent reasoning trajectories, which are represented as a set of candidate graphs $\mathcal{S} = {g_1, g_2, \dots, g_k}$. Each $g_i = (V_i, E_i)$ represents a potential logical reasoning path leading to an answer, where nodes $V$ denote atomic reasoning steps and edges $E$ represent logical dependencies. These candidate graphs are then aggregated and refined to construct high-quality training data for supervising the LLM in learning structured self-graph reasoning.

\noindent\textbf{Graph Integration and Data Cleaning.}
Individual reasoning paths often contain fragmented or suboptimal logic reasoning path. To construct an optimal reasoning trajectory, we integrate all candidate graphs $\mathcal{S}$ into a unified, optimal reasoning graph $\hat{g}$, which we define as the integrated graph that best preserves logical consistency and convergence toward the correct answer. The resulting graph is linearized into a structured template as follows:
\begin{table}[htbp]
\centering
\small
\begin{tabular}{p{0.45\textwidth}}
\toprule
\texttt{<reasoning>} \\
\quad \texttt{<step>} $v_i \to v_j$ \texttt{</step>} \\
\quad \dots \\
\texttt{</reasoning>} \\
\texttt{<answer>} Final Answer \texttt{</answer>} \\
\bottomrule
\end{tabular}
\end{table}

\noindent Furthermore, to ensure the integrity of the training data, we evaluate the final answer $\hat{A}$ derived from the synthesized graph $\hat{g}$ against the known correct answer $L$ for question $Q$. The graph $\hat{g}$ is retained in our final training set $\mathcal{D}$ if and only if the reasoning process leads to the correct answer:
\begin{equation}
\mathcal{D} = \left\{ (Q, \hat{g}, L) \mid f(\hat{g}) = L, \, f: \hat{g} \to \hat{A} \right\}.
\end{equation}
By discarding instances where the synthesized reasoning fails to reach the correct conclusion, we ensure that SGR is trained exclusively on logically consistent and factually grounded reasoning-answer pairs.

\subsection{Self-Graph Reasoning Framework}
The core of SGR is to transform the LLM's latent cognitive process into an explicit, verifiable graph topology before reaching a conclusion.

\paragraph{Supervised Graph Learning.}
To empower the LLM with structured self-reasoning capability, we perform Supervised Fine-Tuning (SFT) with LoRA~\cite{hu2022lora} on the training data $\mathcal{D}$ constructed in Section~\ref{reasoning_graph_construction}. Each training instance is a triplet $(Q, \hat{g}, L)$, where the graph $\hat{g}$ serves as the structural intermediate reasoning between the question $Q$ and the ground truth $L$. Formally, we optimize the model parameters $\theta$ by minimizing the standard cross-entropy loss over the reasoning graph and the final answer:
\begin{equation}
\mathcal{L}(\theta) = - \sum_{(Q, \hat{g}, L) \in \mathcal{D}} \log P_\theta (\hat{g}, L \mid Q).
\end{equation}
By generating $\hat{g}$ token-by-token, the model effectively constructs a structural reasoning graph that constrains the final answer $L$ to be a direct consequence of verified antecedent steps. Compared to \textit{Direct Answering}, which collapses the reasoning space into a single mapping, and \textit{Linear Reasoning} (e.g., CoT), which is restricted to a single-path dependency, SGR ensures logical consistency between the reasoning process and the final answer. Each node $v_j$ must be explicitly justified by its parent nodes $Pa(v_j)$, thereby eliminating the "logical drift" often observed in linear reasoning~\cite{lanham2023measuring}. By externalizing internal thoughts into $V$ and $E$, SGR provides a transparent and structured reasoning process. 

\paragraph{Inference Stage.}
At inference time, given a question $Q$, the model generates a structured reasoning graph $\hat{g}$ followed by the final answer $L$, according to the learned conditional distribution $P_\theta(\hat{g}, L \mid Q)$. An illustrative example is shown in Figure~\ref{fig:sample}. Unlike Direct Answering, which directly outputs an answer without any intermediate reasoning, or Linear Reasoning, which may produce a correct answer but with inconsistent or hallucinatory reasoning (e.g., wrongly stating that \textit{The Sorcerer’s Stone was released in the 1990s}), our Self-Graph Reasoning (SGR) generates a structured reasoning graph that provides a consistent and clear reasoning process. As shown, two reasoning branches respectively consider \textit{“the release date of the first iPhone”} and \textit{“the release date of the first Harry Potter movie,”} which eventually converge to the correct conclusion. Each step is explicit and verifiable, allowing easy detection of possible hallucinations or reasoning errors.

\begin{table*}[t]
    \centering
    \resizebox{1.0\textwidth}{!}{
    \begin{tabular}{lccccccc}
    \toprule
         \multicolumn{1}{c}{} &  \multicolumn{3}{c}{\textbf{General Domain}} && \multicolumn{2}{c}{\textbf{Specialized Domain}} & \multicolumn{1}{c}{} \\
         \cmidrule(lr){2-5} \cmidrule(lr){6-7}
         \textbf{Method} & \textbf{LogiQA} & \textbf{AIW} & \textbf{AR-LSAT} && \textbf{MedQA} & \textbf{MathQA} & \makecell{\textbf{Overall} \\Avg. (\%)} \\
        \midrule
         \textit{\textbf{Proprietary LLMs}} &&&&&&& \\
         GPT-4o~\cite{hurst2024gpt} & 74.01 & 32.50 & 31.75 && 88.29 & 81.05 & 61.52 \\
         GPT-5.1~\cite{gpt5-1} & 76.34 & 57.00 & 33.33 && 89.55 & 39.09 & 59.06\\
         Claude-3.5-Haiku~\cite{claude35haiku} & 65.97 & 2.50 & 29.41 && 76.36 & 79.36 & 50.72 \\
         Gemini-2.5-Pro\textsuperscript{*}~\cite{gemini2-5-pro} & 85.75 & 76.00 & 96.22 && - & 73.10 & -\\
         \midrule
         \textit{\textbf{Open-source LLMs}} &&&&&&& \\
         LLaMA-3.2-3B~\cite{dubey2024llama}  & 41.28 & 1.70 & 20.00 && 49.57 & 29.21 & 28.35 \\
         LLaMA-3.1-8B~\cite{dubey2024llama}  & 49.17 & 5.00 & 26.96 && 55.22 & 29.39 & 33.15\\
         LLaMA-3.3-70B~\cite{dubey2024llama} & 64.01 & \underline{19.50} & 31.30 && 63.55 & 38.09 & \cellcolor{cyan!15}43.29\\
         Qwen2.5-7B~\cite{yang2024qwen2} & 34.10 & 5.00 & 17.39 && 59.54 & 38.29 & 30.96\\
         Qwen2.5-72B~\cite{yang2024qwen2}  & \textbf{76.91} & 5.00 & \textbf{34.78} && \underline{74.42} & \underline{52.60} & \underline{49.00}\\
         \midrule
         \textit{\textbf{Specialized Graph-based Methods}} &&&&&&& \\
         RwG-LLaMA3.1-70B\textsuperscript{\dag}~\cite{han2025reasoning} & 59.13 & 12.00 & 31.73 && - & - & -\\
         RwG-Claude-3-sonnet\textsuperscript{\dag}~\cite{han2025reasoning} & 45.16 & 2.60 & 30.86 && - & - & -\\
         \midrule
         \textit{\textbf{Ours}} &&&&&&& \\
         SGR-Llama3.3-70B & \underline{69.91} & \textbf{57.50} & \underline{31.74} && \textbf{78.81} & \textbf{67.17} & \cellcolor{pink!42}\textbf{61.03} \\
    \bottomrule
    \end{tabular}}
    \caption{Accuracy of models across all benchmarks, covering both general-domain and specialized-domain question answering. Methods are categorized into proprietary LLMs, open-source LLMs, and specialized reasoning approaches. The best results among non-proprietary LLMs are highlighted in \textbf{bold}, and the second-best are \underline{underlined}. The \colorbox{pink!42}{red} and \colorbox{cyan!15}{blue} indicate the average performance of our SGR method and its base model, respectively. \textsuperscript{*}Gemini results are omitted for medical-domain questions due to policy restrictions.\textsuperscript{\dag}RwG results are taken from the original publication due to reproduction issues; therefore, MedQA and MathQA are excluded.}
    \label{tab:main_result}
    \vspace{-3mm}
\end{table*}

\section{Experimental Setup}
\subsection{Datasets}
\noindent\textbf{Training Dataset.} 
To enable the LLM to perform self-graph reasoning before answering questions, we construct a dataset of about 10K samples based on the training subset of the LogiQA dataset~\cite{liu2020logiqa}. LogiQA is a general-domain QA benchmark that involves complex logical reasoning, making it particularly suitable for supervising structured graph-based reasoning. The final 10K samples are obtained after data cleaning and filtering. For each question–answer pair, we use GPT-4o~\cite{hurst2024gpt} to generate an explicit reasoning graph represented as \{reasoning step$_i$→reasoning step$_j$\}, which captures the logical dependencies leading to the correct answer. The resulting dataset is organized in the format of \{\textit{Question}, \textit{Graph Reasoning}, \textit{Label}\}. More information is shown in Appendix~\ref{appendix:our_data}. We randomly split the dataset into training and validation subsets with a ratio of 9:1.

\noindent\textbf{Evaluation Benchmarks.} 
Our work primarily targets general-domain question answering. 
To evaluate the model’s performance in this setting, we adopt several widely used benchmarks, including the LogiQA test set~\cite{liu2020logiqa}, AIW, AIW+~\cite{nezhurina2024alice}, and AR-LSAT~\cite{wang2022lsat}. In addition, to evaluate the cross-domain generalization of our method, which is trained on general-domain data, we further evaluate it on MedQA~\cite{jin2020disease} and MathQA~\cite{amini2019mathqa}. Details of the benchmarks are provided in Appendix~\ref{appendix_benchmark}.

\subsection{Baselines}
We compare our proposed framework with three categories of models, including proprietary LLMs, open-source LLMs, and specialized methods. 

For proprietary LLMs, we include GPT-4o~\cite{hurst2024gpt}, GPT-5.1~\cite{gpt5-1}, Claude-3.5-Haiku~\cite{claude35haiku}, and Gemini-2.5-Pro~\cite{gemini2-5-pro}, which serve as strong closed-source baselines with advanced reasoning capabilities. 
For open-source LLMs, we evaluate a series of models with varying scales, including LLaMA-3.2-3B, LLaMA-3.1-8B, LLaMA-3.1-8B, LLaMA-3.3-70B~\cite{dubey2024llama}, Qwen2.5-7B, Qwen2.5-72B~\cite{yang2024qwen2}.
We also compare our method with the specialized method Reasoning with Graphs (RwG)~\cite{han2025reasoning}, which relies on externally pre-extracted graphs to enhance model reasoning.


\subsection{Implementation Details}
We perform supervised fine-tuning of LLaMA-3.3-70B~\footnote{\url{https://huggingface.co/meta-llama/Llama-3.3-70B-Instruct}} using LoRA~\cite{hu2022lora}, keeping the base model frozen. Models are trained for 1-3 epochs with early stopping (patience=1), using a batch size of 4 and learning rate \(6 \times 10^{-5}\). Gradient accumulation is employed to effectively increase the batch size. Maximum generation length is set to 1024. The checkpoint with the best validation performance is selected for evaluation. Experiments are conducted on 8 NVIDIA A100 40GB GPUs. Full hyperparameter details are provided in the Appendix~\ref{appendix:hyperparameters}. We use accuracy (Acc) as the evaluation metric for our experiments, following previous methods~\cite{cao2024graphreason, han2025reasoning}.

\section{Results and Analysis}
\subsection{Main Results}
Table \ref{tab:main_result} reports the accuracy of various models, including proprietary LLMs, open-source LLMs, and specialized graph-based methods, across general-domain and specialized-domain question answering benchmarks. In general, our SGR-Llama3.3-70B achieves an average accuracy of 61.03\%, demonstrating competitive performance across both general and specialized domains. Through efficient graph reasoning fine-tuning, our method performs on par with the powerful proprietary LLM GPT-4o, while substantially outperforming all comparable open-source models of similar scale.

Specifically, in the general-domain setting, including LogiQA, AIW, and AR-LSAT, our SGR-Llama3.3 70B achieves an average accuracy of 53.05\%, surpassing GPT-4o (46.08 \%) and significantly outperforming all existing open-source LLMs. Compared to its base model LLaMA-3.3-70B, our approach brings a 17.74\% improvement. Particularly on the AIW dataset that relies on logical reasoning, our method achieves 57.50\%, outperforming GPT-4o by 25 points and LLaMA-3.3-70B by 38 points on average across the three benchmarks. Compared with graph-based baselines, our method achieves an 18.76\% higher average accuracy than RwG-LLaMA3.1-70B (52.73\%), which depends on pre-extracted external graphs for reasoning. These results highlight the effectiveness of self-graph reasoning, especially under complex reasoning scenarios in general-domain tasks.

\begin{figure}[t]
\centering
    \includegraphics[width=0.95\linewidth]{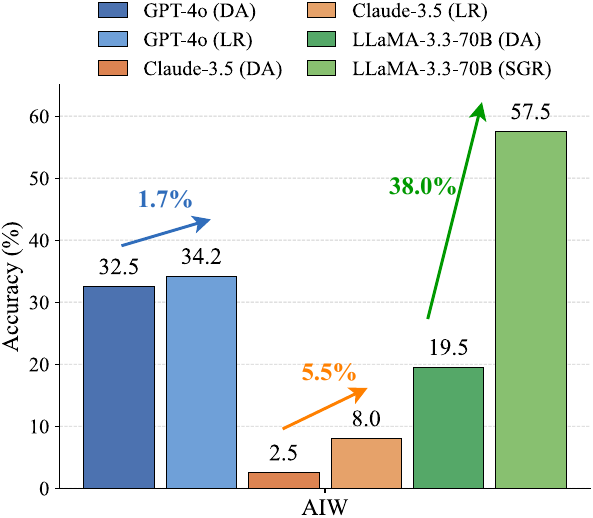}
    \caption{Comparison of accuracy under different reasoning paradigms on the AIW dataset. We evaluate three paradigms: Direct Answering (DA), Linear Reasoning (LR), and our Self-Graph Reasoning (SGR) across GPT-4o, Claude-3.5-haiku, and LLaMA-3.3-70B.
    }
    \label{fig:reasoning_paradigms}
\end{figure}

To further evaluate the generalization capability of our method, we conduct experiments on specialized-domain tasks, including MedQA and MathQA. Notably, the training data are from general-domain sources, containing no domain-specific knowledge of medicine or mathematics. SGR-Llama3.3 70B achieves 78.81\% on MedQA and 67.17\% on MathQA, reaching an average accuracy of 72.99\%, a 22.17\% improvement over the base model. These results indicate that self-graph reasoning enables LLMs to develop a structured, human-like reasoning process that generalizes beyond specific domains and remains effective even without task-specific fine-tuning.

\subsection{Comparison with Existing Paradigms.}
To further demonstrate the effectiveness of our self-graph reasoning (SGR) compared with existing reasoning paradigms, we conduct comparative experiments on the AIW dataset, which requires strong reasoning ability. Specifically, we evaluate three paradigms: direct answering, linear reasoning (CoT), and our SGR approach.
For CoT, we apply standard CoT prompting to both GPT-4o and Claude-3.5-Haiku. For comparison, we train LLaMA-3.3-70B with our SGR framework as the representative implementation of our method. The results are presented in Figure \ref{fig:reasoning_paradigms}.
The experimental results show that, compared to direct answering, GPT-4o achieves a 1.7\% gain and Claude-3.5-Haiku achieves a 5.5\% gain with CoT reasoning. In contrast, our SGR-LLaMA-3.3-70B achieves a remarkable 38\% improvement, surpassing both proprietary models. These results demonstrate the effectiveness of graph-structured reasoning, which enables clearer and logically consistent reasoning processes.

\subsection{Ablation Studies}
\begin{figure}[t]
\centering
    \includegraphics[width=1.0\linewidth]{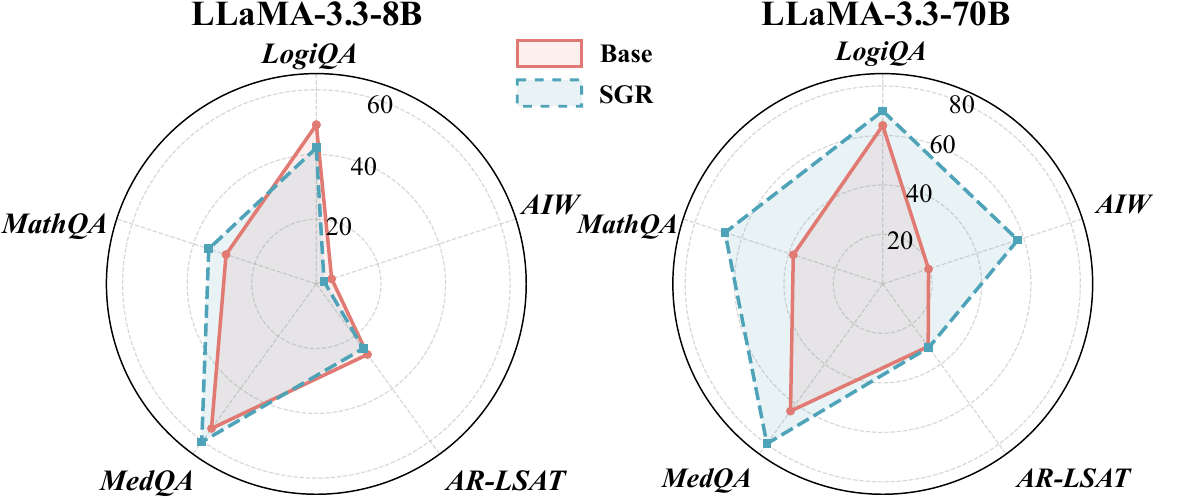}
    \caption{Comparison of accuracy between the base models and our SGR across all five benchmarks for LLaMA-3.3-8B and LLaMA-3.3-70B.
    }
    \label{fig:radar}
    \vspace{-3mm}
\end{figure}

To further assess the effectiveness of our proposed SGR, we conduct ablation studies across LLMs of different scales, comparing base models with and without self-graph reasoning. We evaluate LLaMA-3.3-8B and LLaMA-3.3-70B, with results summarized in Figure~\ref{fig:radar}. The results show that the 8B model exhibits only marginal gains on MathQA and MedQA, and even degradation on more complex general-domain QA datasets, suggesting that limited base capabilities constrain it to effectively perform self-graph reasoning. In contrast, the 70B model demonstrates substantial and consistent improvements across all benchmarks.

Overall, these findings indicate that the benefits of SGR correlate with the model’s underlying reasoning capacity, with stronger LLMs better able to reason in a graph-structured manner, and the impact of SGR becomes more pronounced as model size increases. We expect that applying SGR to even larger or more capable LLMs will yield further improvements. Further experimental analyses are presented in the Appendix.

\subsection{Case studies}
\begin{figure*}[t!]
    \centering
    \includegraphics[width=1\textwidth]{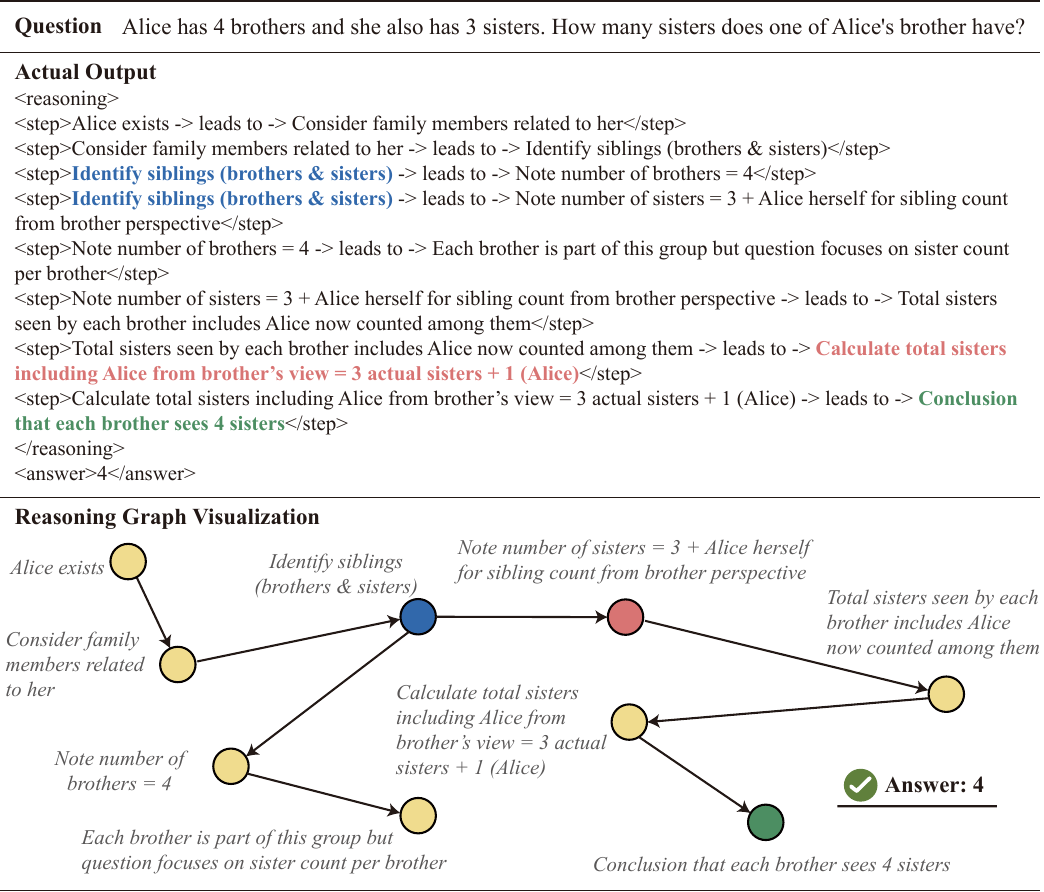}
    \caption{A case study in general-domain QA, illustrating the model’s actual output and its reasoning graph visualization. The blue nodes denote branch nodes, the red nodes denote key nodes, and the green node denotes the final decision node.}
    \label{fig:case_study}
    \vspace{-3mm}
\end{figure*}

We conduct a case study on general-domain QA to illustrate how our method enables LLMs to perform self-graph reasoning. We also visualize the reasoning graph generated by our model, as shown in Figure~\ref{fig:case_study}. Unlike linear reasoning paradigms such as  CoT, SGR explicitly constructs a structured reasoning graph before producing the final answer. This graph representation clearly demonstrates how the model decomposes the problem into interpretable reasoning units. 

For the given question, the model generates a reasoning graph consisting of 9 nodes and 8 directed edges, where each node corresponds to a reasoning step, and each edge encodes an explicit dependency between reasoning units. This structured organization makes the reasoning process clearer and reduces the risk of potential context inconsistency and error propagation that may arise in linear reasoning paradigms. Specifically, the branching node \textit{“Identify siblings (brothers \& sisters)”} separates the interpretation of the two sibling types, creating two independent reasoning branches. This prevents the mixing of different reasoning units, a common issue in linear CoT, and allows the model to focus separately on the brother-related and sister-related subproblems. Notably, in the “brothers” branch, SGR correctly halts further expansion once it recognizes that the question concerns the sister count per brother, and subsequently redirects reasoning to the sister-related branch.

Further, a crucial advantage of SGR is evident in the node \textit{“Note number of sisters = 3 + Alice herself for sibling count from brother perspective”}, which explicitly encodes the perspective shift required to answer the question. The structured reasoning graph enforces the creation of an independent node to handle this logical transition from Alice’s view (3 sisters) to a brother’s view, where Alice must now be counted as a sister. This step is typically implicit or mishandled in linear reasoning paradigms. By explicitly constructing this node, SGR ensures that this key reasoning step is neither overlooked nor weakened.

Finally, SGR aggregates information from multiple upstream nodes through the intermediate computation node \textit{“Calculate total sisters including Alice from brother’s view = 3 actual sisters + 1 (Alice)”}, and converges on the decision node \textit{“Conclusion that each brother sees 4 sisters.”} The graph structure ensures that intermediate reasoning aligns with the final output, leading the model to the \underline{correct answer: 4}.

\section{Conclusion}
In this work, we presented Self-Graph Reasoning (SGR), a framework that enables LLMs to perform graph-structured reasoning on their own for open-domain question answering. SGR overcomes the limitations of linear reasoning and externally provided graphs by allowing models to express their reasoning explicitly in a structured form. We also construct a graph-structured reasoning dataset for training LLMs to perform graph-based reasoning effectively. Experiments across five QA benchmarks show that SGR substantially improves reasoning consistency, yielding a 17.74\% average gain over its base model across all benchmarks.

\section*{Limitations}
\noindent\textbf{Training Data.} The scale of the training data for self-graph reasoning is relatively limited. The constructed self-graph reasoning dataset contains approximately 10K training instances, which may constrain the full potential of the proposed SGR framework. We expect that expanding the training data to a larger scale would further improve both the performance and generalization of SGR.

\noindent\textbf{Base Model Scale.} As discussed in the paper, the effectiveness of SGR is inherently tied to the capability of the base models. Due to computational constraints, our experiments are conducted on a 70B model. Applying SGR to larger-scale models may further improve reasoning consistency and overall performance.



\bibliography{custom}

\clearpage
\appendix
\label{sec:appendix}
\section{Benchmark Details}
\label{appendix_benchmark}
\subsection{General Domain Benchmarks}
\noindent{\textbf{LogiQA.}} The LogiQA dataset~\cite{liu2020logiqa} is a multiple-choice reading comprehension benchmark designed to evaluate logical reasoning in natural language understanding. Each instance consists of a context, a question, and four candidate answers, only one of which is logically correct. The questions are sourced from real-world examination materials, such as national civil service and postgraduate entrance tests, ensuring a high level of reasoning complexity. We use the training set of LogiQA2.0 to construct our graph reasoning dataset and the test set as a benchmark for evaluation.

\noindent{\textbf{AIW.}} The AIW dataset~\cite{nezhurina2024alice} is a benchmark of reasoning problems. Each problem typically presents a simple natural language scenario, such as familial relationships with variables, and requires the model to derive a logically correct answer from basic commonsense logic. Unlike traditional multiple-choice datasets, AIW focuses on minimal text examples that expose reasoning breakdowns in state-of-the-art models, making it a useful diagnostic for evaluating core inferential competence.

\begin{table}[b]
    \centering
    \resizebox{0.48\textwidth}{!}{
    \begin{tabular}{lcccc}
        \toprule
        Dataset & Size & Question Length\\
        \midrule
        \textit{\textbf{General Domain}} &&&& \\
        LogiQA & 1572 & 930.57\\
        AIW  & 200 & 98.79\\
        AR-LSAT & 230 & 892.36\\
        \midrule
        \textit{\textbf{Specialized Domain}} &&&& \\
        MedQA  & 1273 & 877.57 \\
        MathQA  & 2985 & 216.76 \\
        \bottomrule
    \end{tabular}}
    \caption{Statistics of Benchmark Datasets. We report the size of each benchmark and the average text length of questions.}
    \label{tab:datasets}
\end{table}

\noindent{\textbf{AR-LSAT.}} The AR-LSAT dataset~\cite{wang2022lsat} is a large-scale benchmark constructed from the analytical reasoning (logic games) section of the official LSAT (Law School Admission Test). It is designed to evaluate the formal reasoning and deductive inference abilities of language models. Each problem describes a context and requires the model to select the correct answer. We use the test set as a benchmark for evaluation.

\subsection{Specialized Domain Benchmarks}
\noindent{\textbf{MedQA.}} The MedQA dataset~\cite{jin2020disease} is a large-scale medical question answering benchmark designed to evaluate the clinical knowledge and reasoning abilities of language models. Each question is a multiple-choice item sourced from real-world medical licensing examinations. The dataset covers a wide range of medical domains, requiring models to integrate factual recall with domain-specific reasoning.

\noindent{\textbf{MathQA.}} The MathQA dataset~\cite{amini2019mathqa} is a large-scale benchmark for evaluating the mathematical reasoning and problem-solving abilities of language models. It is gathered by using a new representation language to annotate over the AQuA-RAT dataset, covering a wide range of mathematical domains. Each question requires the model to translate a natural language description into a formal reasoning process and compute the correct numerical answer.

\begin{table}[t]
    \centering
    \resizebox{0.40\textwidth}{!}{
    \begin{tabular}{lc}
    \toprule
         Hyperparameter & Value\\
         \midrule
         seed & 42 \\
         batch\_size & 8 \\
         num\_epochs & 1--3 \\
         learning\_rate & $5\times 10^{-6}$ \\
         weight\_decay  & 0.01 \\
         grad\_steps & 4 \\
         warmup & 0.05 \\
         early\_stop\_patience & 1\\
         \midrule
         lora\_r & 8 \\
         lora\_alpha & 16 \\
         lora\_dropout & 0.1 \\
         lora\_target\_modules & q\_proj, v\_proj \\
         \midrule
         max\_txt\_len & 1024 \\
         max\_new\_tokens & 1024 \\
    \bottomrule
    \end{tabular}}
    \caption{Hyperparameters.}
    \label{tab:hyperparameters}
\end{table}

\begin{table*}[t]
    \centering
    \resizebox{1.0\textwidth}{!}{
    \begin{tabular}{lccccccc}
    \toprule
         \multicolumn{1}{c}{} &  \multicolumn{3}{c}{\textbf{General Domain}} && \multicolumn{2}{c}{\textbf{Specialized Domain}} & \multicolumn{1}{c}{} \\
         \cmidrule(lr){2-5} \cmidrule(lr){6-7}
         \textbf{Method} & \textbf{LogiQA} & \textbf{AIW} & \textbf{AR-LSAT} && \textbf{MedQA} & \textbf{MathQA} & \makecell{\textbf{Overall} \\Avg. (\%)} \\
        \midrule
         LLaMA-3.1-8B & 49.17 & 5.0 & 26.96 && 55.22 & 29.39 & 33.15 \\
         SGR-LLaMA-3.1-8B & 42.05 & 2.50 & 24.68 && 60.36 & 35.03 & 32.92\\
         SGR-LLaMA-3.1-8B w/ GRPO & 56.68 & 5.5 & 24.35 && 59.47 & 36.25 & 36.45 \\
    \bottomrule
    \end{tabular}}
    \caption{Accuracy of the base LLaMA-3.1-8B model, our SRG version, and the GRPO-fine-tuned version across all benchmarks.}
    \label{tab:grpo_result}
\end{table*}

\subsection{Preprocessing for Benchmarks}
To facilitate a unified evaluation across different benchmarks, we standardize all datasets into a consistent \{question, label\} format. For datasets originally containing a separate \textit{context} field, we concatenate them with the question text to form a single \textit{question} input. The details are summarized in Table~\ref{tab:datasets}.

\begin{table}[b]
    \centering
    \resizebox{0.4\textwidth}{!}{
    \begin{tabular}{lc}
        \toprule
        Model & Cost (\$) \\
        \midrule
        GPT-4o & \textasciitilde26 \\
        Claude-3.5-haiku   & \textasciitilde10 \\
        Gemini-2.5-Pro  & \textasciitilde24 \\
        GPT-4o CoT      & \textasciitilde80 \\
        Claude-3.5-haiku CoT   & \textasciitilde32 \\
        \midrule
        SGR-Llama3.3-70B(Ours)  & \textasciitilde33.6 \\
        \bottomrule
    \end{tabular}}
    \caption{Comparison of the cost of our method with other LLMs on the LogiQA test set.}
    \label{tab:cost}
\end{table}

\section{Our Graph Reasoning Dataset}
\label{appendix:our_data}
\begin{figure*}[t]
    \centering
    \includegraphics[width=1\linewidth]{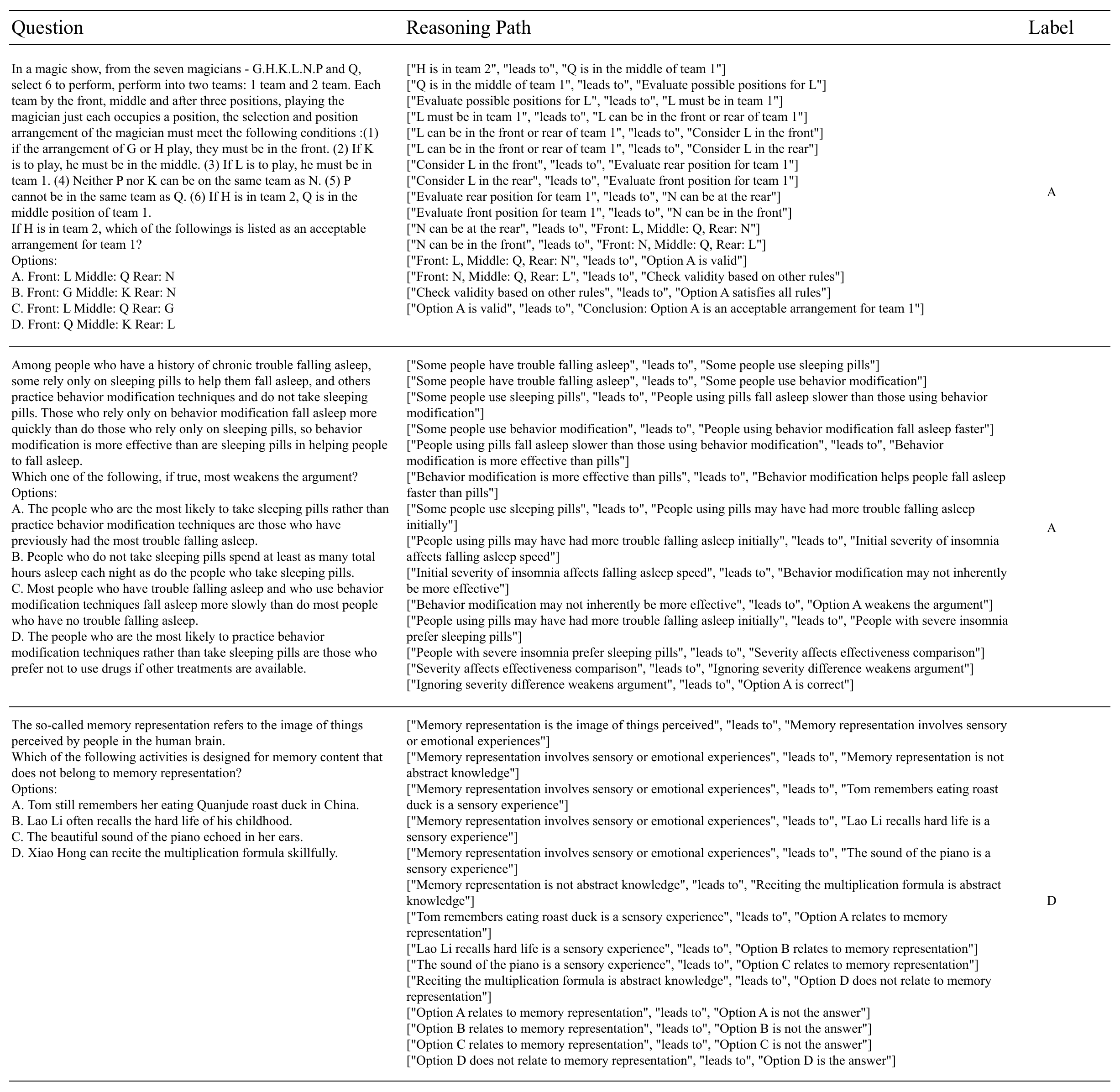}
    \caption{Samples of our graph reasoning data. Each sample consists of a question, its corresponding reasoning graph, and the correct answer (label).}
    \label{fig:data_samples}
\end{figure*}

We construct our graph-based reasoning dataset based on the training set of LogiQA 2.0. The original LogiQA 2.0 training corpus contains a total of 12,547 samples. After reasoning graph construction and data cleaning, we retain 9,869 high-quality samples for our graph reasoning benchmark. Examples of the constructed graph reasoning data is shown in Figure~\ref{fig:data_samples}.

\section{Detail of Hyperparameter.}
\label{appendix:hyperparameters}
We list all the parameters used for SGR-Llama3.3 70B, as shown in table \ref{tab:hyperparameters}. This includes configuration details such as batch size, learning rate, LoRA, and optimizer settings.

\section{Analysis of Computational Cost}
We compared the computational cost of our proposed SRG-LLaMA-3-70B model with several LLMs on the LogiQA benchmark. For our locally deployed model, the cost was estimated at \$0.8 per GPU hour, requiring approximately 42 GPU hours in total. As shown in Table~\ref{tab:cost}, our method incurs a total cost of about \$33.6, which is substantially lower than that of GPT-4o (CoT) (approximately \$80), while achieving comparable performance and offering a more interpretable, graph-structured reasoning process. Given that our model can be deployed locally with a moderate computational budget, these results highlight the efficiency and scalability of the SRG framework for reasoning tasks.

\section{GRPO Fine-Tuning}
To mitigate the potential dilution of supervision on the final answer caused by intermediate reasoning steps, we perform an additional round of GRPO fine-tuning following standard SFT. This stage strengthens both the structured reasoning format and the accuracy of the final answer through two complementary reward functions. The first reward evaluates whether the model’s output strictly adheres to the predefined structured reasoning–answer template, while the second reward assesses whether the content within the \textit{<answer>} tag matches the label $y_i$

\begin{table}[b]
    \centering
    \resizebox{0.5\textwidth}{!}{
    \begin{tabular}{lccccccc}
    \toprule
         \textbf{Method} & \textbf{LogiQA} & \textbf{AIW} & \textbf{MedQA} \\
        \midrule
         Qwen3-8B & 73.20 & 43.00 & 66.80 \\
         Qwen3-8B-thinking & 80.40 & 80.00 & 75.80 \\
         GPT-5.1 & 76.34 & 57.00 & 89.55 \\
         SGR-LLaMA-3.3-70B & 69.91 & 57.50 & 79.81 \\
    \bottomrule
    \end{tabular}}
    \caption{Comparison of accuracy for our Self-Graph Reasoning (SGR) framework against recent reasoning LLMs, including Qwen-3-8B-thinking and other LLMs, on the LogiQA, AIW, and MedQA datasets.}
    \label{tab:reasoningLLM}
\end{table}

We conduct GRPO fine-tuning based on the LLaMA-3.3-8B, using a batch size of 8, a maximum of 6000 steps, and a learning rate of $5\times 10^{-6}$. This post-SFT training phase encourages the model to preserve structured reasoning while maximizing the correctness of the final answer. Results in Table~\ref{tab:grpo_result} indicate that the additional GRPO fine-tuning substantially enhances the model’s answer accuracy, particularly on the in-domain LogiQA dataset, where it achieves a 7.51\% improvement over the base model. This demonstrates that GRPO effectively mitigates the performance degradation associated with the limited capacity of small-scale LLMs. However, we also observe that the reasoning process of the GRPO-tuned model becomes overly concise, occasionally omitting intermediate reasoning steps. These findings suggest an inherent trade-off between answer accuracy and reasoning completeness in small-scale language models.

\section{Comparison with Reasoning LLMs}

To assess the effectiveness of our Self-Graph Reasoning (SGR) framework, we compare its performance against the recent reasoning LLM Qwen-3-8B-thinking~\cite{yang2025qwen3}, evaluated on the LogiQA, AIW, and MedQA datasets, as shown in Table~\ref{tab:reasoningLLM}. Qwen-3B-thinking achieves strong results across all three benchmarks, even outperforming the proprietary GPT-5.1 model. This may be partly due to the inclusion of these public benchmarks in its pretraining or instruction-tuning data, which likely provides prior exposure to the evaluation distributions. In contrast, our SGR-LLaMA-3.3-70B model is fine-tuned with only a small amount of graph data and without any benchmark-specific priors, yet it still achieves competitive accuracy.

\section{Prompts}
\label{sec:prompts}
\begin{figure*}[t]
    \centering
    \includegraphics[width=1\linewidth]{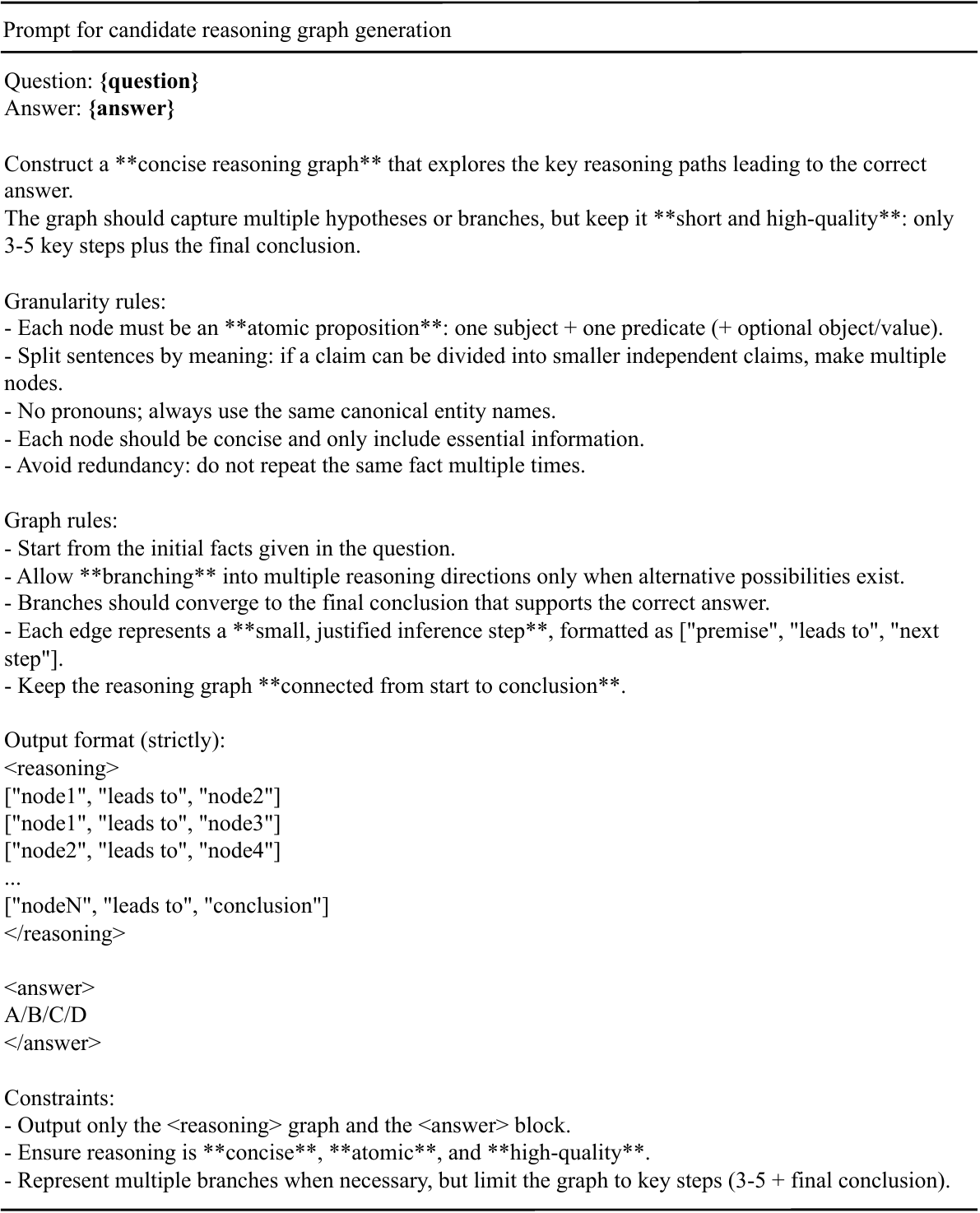}
    \caption{Prompt for candidate reasoning graph generation.}
    \label{fig:prompt_graph1}
\end{figure*}

\begin{figure*}[t]
    \centering
    \includegraphics[width=1\linewidth]{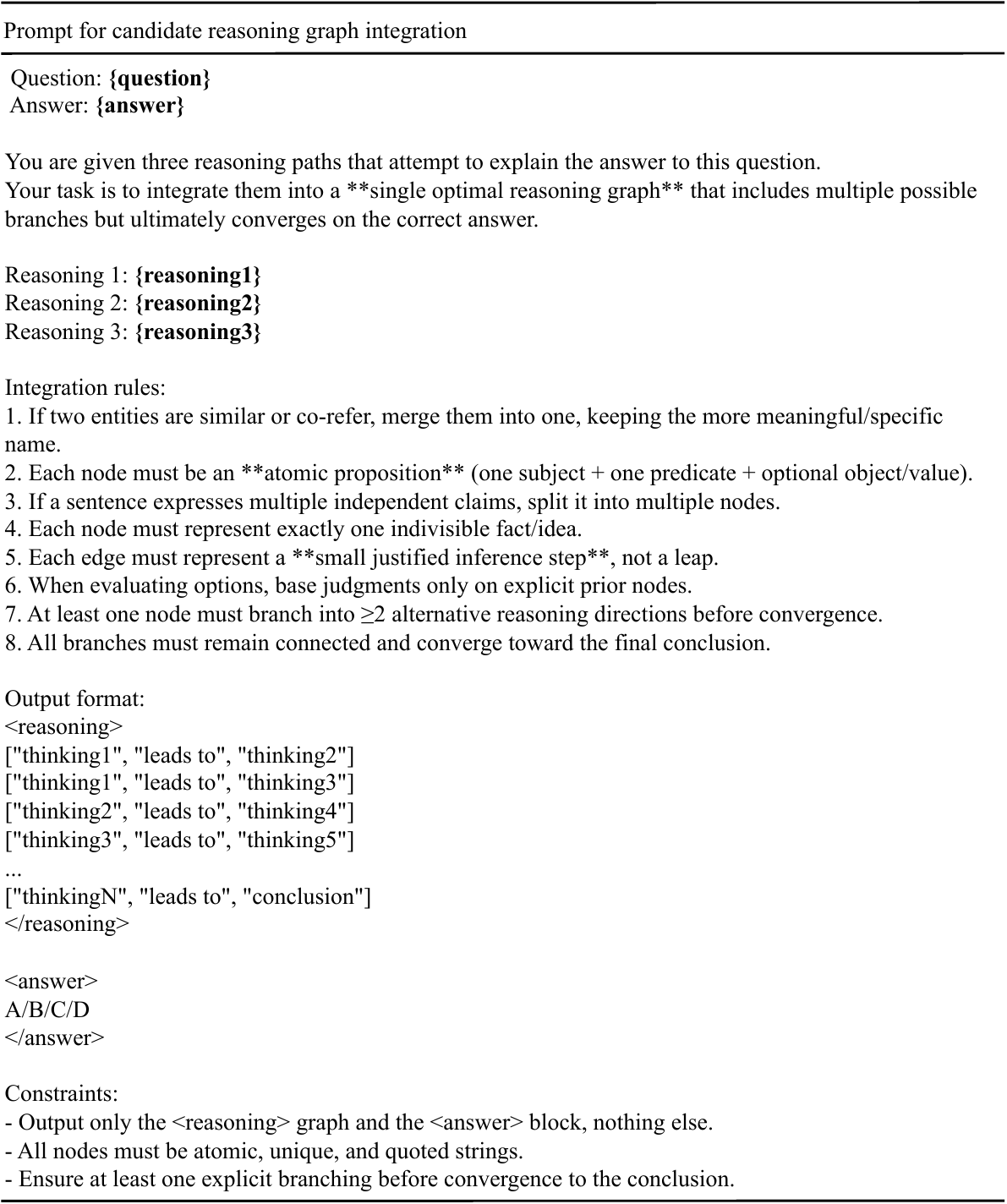}
    \caption{Prompt for candidate reasoning graph integration.}
    \label{fig:prompt_graph2}
\end{figure*}

\begin{figure*}[t]
    \centering
    \includegraphics[width=1\linewidth]{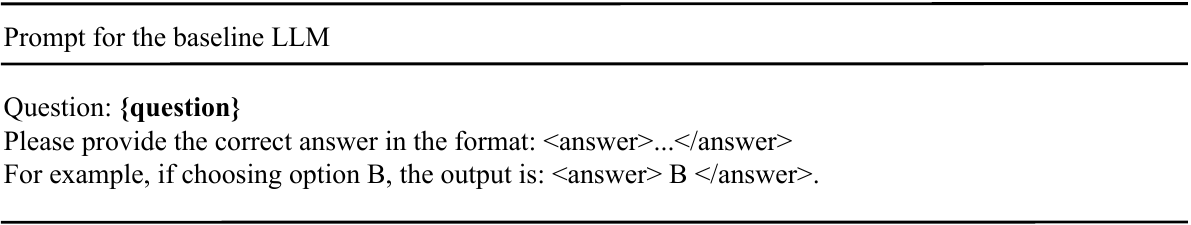}
    \caption{Prompt for the baseline LLMs.}
    \label{fig:prompt1}
\end{figure*}

\begin{figure*}[t]
    \centering
    \includegraphics[width=1\linewidth]{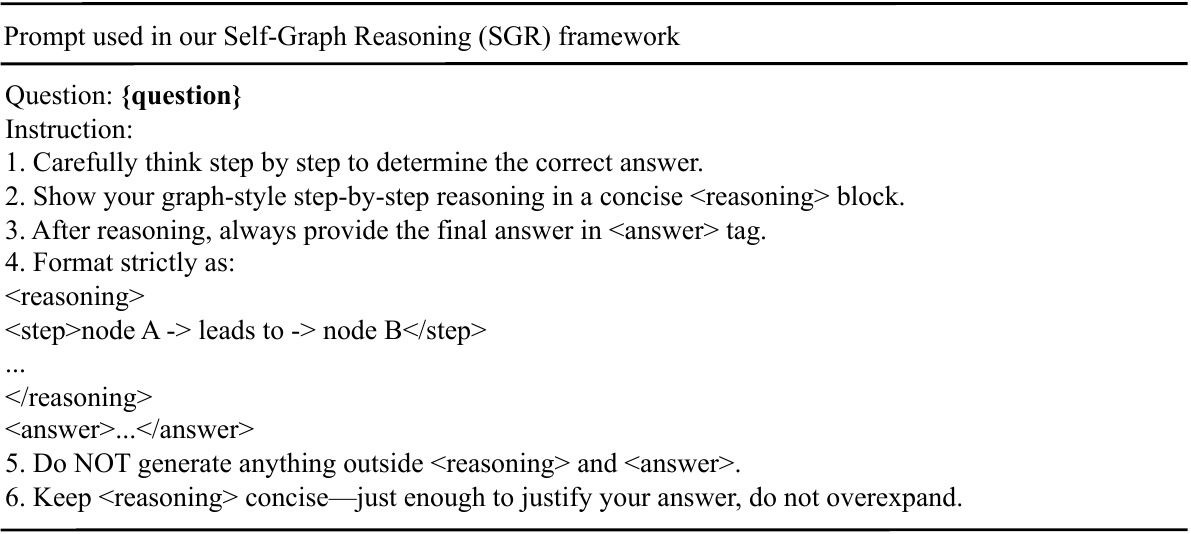}
    \caption{Prompt used in our Self-Graph Reasoning (SGR) framework.}
    \label{fig:prompt2}
\end{figure*}

We provide the prompts used in our experiments.  
To generate candidate reasoning graphs, the model is prompted as illustrated in Figure~\ref{fig:prompt_graph1}.  
For integrating candidate graphs into an optimal reasoning graph, the model is prompted as shown in Figure~\ref{fig:prompt_graph2}.  
A standard question-answering prompt is constructed for the baseline LLMs, as depicted in Figure~\ref{fig:prompt1}.  
Within our Self-Graph Reasoning (SGR) framework, the model is prompted as illustrated in Figure~\ref{fig:prompt2}.

\end{document}